# Is Multilingual BERT Fluent in Language Generation?

**Samuel Rönnqvist***   **Jenna Kanerva***   **Tapio Salakoski**   **Filip Ginter***
TurkuNLP
Department of Future Technologies
University of Turku, Finland
{saanro,jmnybl,sala,figint}@utu.fi

## Abstract

The multilingual BERT model is trained on 104 languages and meant to serve as a universal language model and tool for encoding sentences. We explore how well the model performs on several languages across several tasks: a diagnostic classification probing the embeddings for a particular syntactic property, a cloze task testing the language modelling ability to fill in gaps in a sentence, and a natural language generation task testing for the ability to produce coherent text fitting a given context. We find that the currently available multilingual BERT model is clearly inferior to the monolingual counterparts, and cannot in many cases serve as a substitute for a well-trained monolingual model. We find that the English and German models perform well at generation, whereas the multilingual model is lacking, in particular, for Nordic languages.[1]

## 1 Introduction

The language representation model BERT (Bidirectional Encoder Representations from Transformers) has been shown to achieve state-of-the-art performance when fine-tuned on a range of downstream tasks related to language understanding (Devlin et al., 2018), and recently also language generation. In addition to downstream applications, many recent studies have explored more directly how various types of linguistic information is captured in BERT's representations.

However, all such studies we are aware of are conducted for English using the monolingual BERT model as the availability of pretrained BERT models for other languages is extremely scarce. For the vast majority of languages, the only option is the multilingual BERT model trained jointly on 104 languages. In "coffee break" discussions, it is often mentioned that the multilingual BERT model lags behind the monolingual models in terms of quality and cannot serve as a drop-in replacement.

In this paper, we therefore set out to test the multilingual model on several tasks and several languages (primarily Nordic), to establish whether, and to what extent this is the case, as well as to establish at least an order-of-magnitude expectation of the performance of the present multilingual BERT model on these tasks and languages. It must be stressed that this paper deals with the particular multilingual model distributed by the BERT creators, rather than the more general question of comparison of the multilingual and monolingual training schedule. Studying those questions would necessitate training multilingual BERT models with resource requirements far beyond those at our disposal.

We put a particular focus on the natural language generation (NLG) task, which we hypothesize requires a deeper understanding of the language in question on the side of the model. We take English and German, for which monolingual versions of BERT are available, as reference languages, in order to compare how they perform in the mono- vs. multilingual settings. Furthermore, we perform experiments with the Nordic languages of Danish, Finnish, Norwegian (Bokmål and Nynorsk) and Swedish, with in-depth evaluations on Finnish and Swedish, as well as the abovementioned two reference languages.

## 2 Related Work

A BERT model is comprised of several layers of stacked Transformer networks (Vaswani et al.,

---

*The marked authors contributed equally to this paper.
[1] The code of the experiments in the paper is available at: https://github.com/TurkuNLP/bert-eval

2017), each providing representations of both the input sequence and its individual tokens. The model incorporates a tokenizer that splits an input sentence into words, or subword units for words or word forms that are relatively infrequent in the training data.

Several recent studies have explored how BERT captures linguistic information in English, and how it is distributed across layers. (Tenney et al., 2019; Jawahar et al., 2019; Clark et al., 2019) A particular line of inquiry has focused on how much hierarchical understanding of a language and knowledge of the syntactic structure is captured in the word representations of the monolingual English BERT model. In Goldberg (2019), the BERT models are shown to perform well on capturing several syntactic phenomena of the English language. The paper shows the model to favor the correct subject-verb agreement over the wrong one even if the input is crafted to mislead the model with agreement attractors, i.e. an intervening subordinate clause with opposite number of the subject. BERT is also shown to perform well on the agreement task even if tokens are randomly substituted from the same part-of-speech category, making the input semantically meaningless while preserving the syntactic structure.

Similarly, Ettinger (2019) evaluates the BERT model on several English psycholinguistic datasets, where the model is shown generally being able to distinguish a good completion from a bad one, while still failing in some more complex categories, for example being insensitive to negation.

Lin et al. (2019) uses a diagnostic classifier to study to which extent syntactic or positional information can be predicted from the English BERT embeddings, and how this information is carried through the different layers.

The multilingual BERT model is studied in the context of zero-shot cross-lingual transfer, where it is shown to perform competitively to other transfer models. (Pires et al., 2019; Wu and Dredze, 2019)

Text generation with BERT is introduced by Wang and Cho (2019), who demonstrate several different algorithms to generate language with a BERT model. They demonstrate that BERT even though not being trained on an explicit language generation objective, is capable of generating coherent, varied language.

| Language | BERT | Test acc. | Baseline |
|---|---|---|---|
| English | mono | 86.03 | 54.93 |
|  | multi | 87.82 | 54.44 |
| German | mono | 97.27 | 69.61 |
|  | multi | 95.29 | 69.19 |
| Danish | multi | 89.96 | 53.25 |
| Finnish | multi | 93.20 | 50.54 |
| Nor. (Bokmål) | multi | 93.67 | 56.19 |
| Nor. (Nynorsk) | multi | 94.44 | 53.18 |
| Swedish | multi | 93.00 | 62.09 |

Table 1: Diagnostic classifier results. Auxiliary classification task accuracies and majority class baselines for all languages.

## 3 Experiments

We evaluate the BERT models on 6 languages, English, German, Swedish, Finnish, Danish, and Norwegian (Bokmål and Nynorsk), and three different tasks. In addition to automatic metrics, the generated output is manually evaluated for English, German, Swedish, and Finnish, the four languages that at least one of the authors is fluent in, and therefore comfortable evaluating. For English and German there are monolingual BERT models available, which we use as references to evaluate the performance of the multilingual BERT model.[2] We further compare performance among these languages and the four Nordic languages in order to assess its utility for such relatively low-resource languages. In all evaluation tasks, we use data from the Universal Dependencies (UD) ver 2.4 treebanks (Nivre et al., 2016, 2019) for the languages in question.[3]

### 3.1 Diagnostic Classifier

As an initial experiment, we train a diagnostic classifier to predict whether an auxiliary is the main auxiliary of its sentence, in order to assess how well the BERT encodings represent elementary linguistic information including hierarchical understanding of a sentence. The task is inspired by Lin et al. (2019) who use it as one way of testing what kind of linguistic knowledge BERT is able to encode. Specifically, they use it as a proxy for assessing whether BERT has a hierarchical representation of sentences, as it is necessary information for differentiating between main and subordinate clause or coordinate clause auxiliaries.

---

[2] For multilingual and English monolingual experiments we used the official models by the original BERT authors, namely `bert-base-multilingual-cased` and `bert-base-uncased`. For German monolingual experiments we use the model provided by Deepset (`bert-base-german-cased`).

[3] Treebanks are English-EWT, German-HDT (part a), Swedish-Talbanken, Finnish-TDT, Danish-DDT, Norwegian-Bokmaal, and Norwegian-Nynorsk.

All words marked with the part-of-speech tag `AUX` in the treebank data are taken as prediction candidates, where the target is a binary classification as to whether the auxiliary is dependent on the root token of the sentence or not. The input of the classifier is the final-layer BERT embedding for the auxiliary. In case the auxiliary token is tokenized into multiple subword units, each subword representation is fed as a separate instance, and thus classified independently. We expect each subword embedding to encode the relevant knowledge of both the whole word and its function in the sentence.

The classifier consists of 768 input units corresponding to the BERT base embedding size and a softmax layer. The model is trained separately for all languages and available BERT model configurations, using treebank training sections, and SGD otimizer for 50 epochs. For improved comparability, the train set size is capped for all languages to that of the smallest treebank (Swedish), for which we were able to extract 3031 training examples. The treebank test sets yield 1002–1217 examples, with the exception of Danish with 515 examples.

The results evaluated on the treebank test sets are listed in Table 1, where we measure subword classification accuracy. The majority class baseline frequencies are listed as reference; they tend to be relatively balanced, although somewhat tilted towards main auxiliaries. There is a notable 2 percentage point decrease for German with multilingual BERT, whereas English exhibits a 1.8 point increase. Comparison between languages is problematic, but we observe that all perform relatively well on the task.[4] Albeit our results not being directly comparable with Lin et al., our findings are in line with their work, indicating BERT being able to encode hierarchical sentence information in all languages, and most interestingly, the same holds also for the multilingual BERT model.

---

[4] The slight variation in baseline between models for the same language is likely influenced by differing tokenization.

|  | Mono | Multi |
|---|---|---|
| English | **45.92** | 33.94 |
| German | **43.93** | 28.10 |
| Swedish |  | 22.30 |
| Finnish |  | 14.56 |
| Danish |  | 25.07 |
| Norwegian (Bokmål) |  | 25.21 |
| Norwegian (Nynorsk) |  | 22.28 |

Table 2: Results for the cloze test in terms of subword predictions accuracy.

## 3.2 Cloze Test

Moving towards natural language generation, and to evaluate the BERT models with respect to their original training objective, we employ a cloze test, where words are randomly masked and predicted back. We mask a random 15% of words in each sentence, and, in case a word is composed of several subwords, all subwords are masked for an easier and more meaningful evaluation. All masked positions are predicted at once in the same manner as done in the BERT pretraining (i.e. no iterative prediction of one position per time step). As a source of sentences, we use the training sections of the treebanks, limited to sentences of 5–50 tokens in length.

The results are shown in Table 2, where we measure subword level prediction accuracy, i.e. how many times the model gives the highest confidence score for the original subword. Overall, we find that the multilingual model substantially lags behind the monolingual variants (at 15–34% vs. 44–45% accuracy), even though the performance at worst is far from trivial. We also observe a notable difference in performance of the multilingual model across the languages, being able to correctly predict between 15% and 34% of the masked subwords. English and German score highest also in the multilingual setting, whereas the Scandinavian languages perform somewhat worse, but similarly among themselves. Finnish stands out as the most challenging.

In order to gain a better understanding of the predictions, we perform a manual evaluation on four languages to observe whether the model is able to fill the gaps with plausible predictions although differing from the original. We manually categorize each predicted word into one of the following categories:

|     |       | match | mismatch | copy | gibb |
|-----|-------|-------|----------|------|------|
| Eng | mono  | **88%** | 9%  | 1% | 1%  |
|     | multi | **72%** | 15% | 8% | 6%  |
| Ger | mono  | **82%** | 12% | 1% | 5%  |
|     | multi | **69%** | 15% | 6% | 10% |
| Fin | multi | **42%** | 15% | 3% | 39% |
| Swe | multi | **56%** | 19% | 2% | 23% |

Table 3: Manual evaluation of words generated in the cloze test.

- **match:** A real word fitting the context both grammatically and semantically

- **mismatch:** A real word that does not fit the context

- **copy:** An unnatural repetition of a word appearing in the nearby context

- **gibberish:** Subwords do not form a real word, or the prediction forms a meaningless sequence of tokens (e.g. sequence of punctuation tokens)

An example prediction of each category is given in Figure 1 and the evaluation results are summarized in Table 3. These even further demonstrate the capability of the monolingual models, with 82–88% of the generated words fitting the context both syntactically and semantically, i.e. being an acceptable substitution for the masked word in the given context.

By contrast, the matches decrease for German and English, using the multilingual model, to 69% and 72% respectively. Finnish and Swedish perform significantly worse, with match rates at 42% and 56%. The evaluation is based on 50–100 sentences per language and model, and about 100–200 predicted words in each case.

The other categories display similar trends: the semantically or syntactically mismatching words increase for the multilingual model, and in particular the amount of gibberish surges for the Nordic languages. The fact that prediction in Finnish exhibits almost twice as much gibberish as in Swedish is likely influenced by the morphological richness of Finnish, resulting in words to generally be composed of more subword units and the likelihood of predicting non-existent words being higher. An interesting trend for the two Nordic languages, especially strongly seen in Finnish, is the predictions mostly falling into two distinct ends of the evaluation scale, 42% being perfectly acceptable substitutions, while 39% being gibberish. The likely explanation noticed during manual evaluation is the model being quite capable predicting natural output in the place of masked function words, while completely failing to predict anything reasonable for masked content words forming longer subword sequences.

|     |       | on-top | off-top | copy | gibb |
|-----|-------|--------|---------|------|------|
| Eng | mono  | **50%** | 21% | 5%  | 24% |
|     | multi | 7%  | 2%  | 38% | **53%** |
| Ger | mono  | **67%** | 28% | 3%  | 2%  |
|     | multi | 17% | 13% | **48%** | 22% |
| Fin | multi | 19% | 2%  | 37% | **43%** |
| Swe | multi | 10% | 5%  | **47%** | 37% |

Table 4: Manual evaluation of generated text from the mono- and multilingual models. The categories are, in order, on-topic original text, off-topic original text, copy of the context, and gibberish. N is 55–60 for all tests.

Examples of the model predictions in this task are given in Figure 2 for English, German, Swedish and Finnish, generated using both monolingual and multilingual models.

## 4 Sentence Generation

To evaluate and compare the text generation abilities of the models, we employ the method recently introduced by Wang and Cho (2019) which enables BERT to be used for text generation.[5] In particular, we use the Gibbs-sampling-based method, reported in the paper to give the best results. In this method, a sequence of [MASK] symbols is generated and BERT is used for a number of iterations to generate new subwords at random individual positions of this sequence until the maximum number of iterations (500 by default), or convergence are reached. This method is shown by Wang and Cho to produce varied output of good quality, even though not entirely competitive with the famous GPT-2 model (Radford et al., 2019). Most importantly for our objective, this method allows us to probe the model's ability to generate longer sequences of the language and to compare the relative differences between the monolingual and multilingual pre-trained BERT models.

---

[5]Note that some of the underlying assumptions of this paper were later corrected by the authors http://tiny.cc/cho-correction

| Labels | Generated |
|---|---|
| match | Question[**ing**∼**about**] the sinking of the Titanic? |
| mismatch | Those [**they**∼**ones**] are quite small. |
| match, copy | I [**felt**∼**understand**] that it is a [**process**∼**competitive**] process... |
| gibberish | A full [**- of and**∼**substantive**] reconciliation of cash and funding accounts |

Figure 1: Example generation of each category used in the manual evaluation of Cloze task predictions. Examples are generated by the English multilingual model. The format of the masked words is [predicted∼gold]. Examples for other languages and models are shown in Figure 2.

| Lang | Model | Generation |
|---|---|---|
| Eng | mono | regarding [**the**∼**those**] rumors about [**people**∼**wolves**] living in yellowstone prior to the official reintroduction? |
| | multi | Regarding [**the**∼**those**] rumors about [**thes**∼**wolves**] living in Yellowstone prior to the official reintroduction? |
| | mono | we [**went**∼**got**] to [**work**∼**talking**] and he got me set up and i [**just**∼**test**] drove with craig and i fell head over heels for this car [**and**∼**all**] i kept saying, "[**but**∼**was**] i gotta have it [**.**∼**!**]" |
| | multi | We [**went**∼**got**] to [**Craig**∼**talking**] and he got me set up and I [**went**∼**test**] drove with Craig and I fell head over heels for this car [**and**∼**all**] I kept saying, "[**And**∼**was**] I gotta have it [**.**∼**!**]" |
| Ger | mono | [**Voraussetzung**∼**Kennzeichen**] für eine [**intensivere**∼**krankhafte**] Nutzung des Internets sei unter anderem ein deutlicher Rückzug [**aus**∼**aus**] dem sozialen Leben. |
| | multi | [**Ein Vorsetzung**∼**Kennzeichen**] für eine [**gewise**∼**krankhafte**] Nutzung des Internets sei unter anderem ein deutlicher Rückzug [**aus**∼**aus**] dem sozialen Leben. |
| | mono | "[**Es**∼**Das**] ist eine Revolution für die mobile Kommunikation", meint [**Professor**∼**Vizepräsident**] Mike Zafirovski. |
| | multi | "[**Es**∼**Das**] ist eine Revolution für die mobile Kommunikation", meint [**der -er**∼**Vizepräsident**] Mike Zafirovski. |
| Fin | multi | Stokessa Gallagher [**oli**∼**pelasi**] enimmäkseen laiturina, joka ei ollut hänen [**valäaäa**∼**lempipaikkojaan**]. |
| | multi | Nykyhetki laajenee vauhdilla, joka [**johtaa**∼**saa**] tulevaisuuden kutistumaan lähes menneisyyden kaltaiseksi [**. .ksi . . , ,**∼**makrokääpiöksi**] [**joka**∼**joka**] vierittää kvarkkia alas leskenlehden terää salaiseen maailmaansa. |
| Swe | multi | Men du [**måste**∼**kan**] få ett givande grepp på staden [**från**∼**och**] dess [**, och**∼**miljö**] också från andra utgångspunkter. |
| | multi | År 1951 [**stod**∼**gjorde**] den engelske [**psnologen .**∼**läkaren**] J. Bowlby för WHO:s [**forsknings för en**∼**räkning**] en sammanställning av dittills gjorda undersökningar över hur [**barnär barn**∼**späda**] och små barn, som för någon tid helt skilts [**från**∼**från**] sin mor, utvecklas. |

Figure 2: Example generations of the cloze prediction task for English, German, Finnish and Swedish. The format of the masked words is [predicted∼gold].

| Model | Judgement | Generated text in context |
|---|---|---|
| Mono | on-topic | It came out better than I even imagined . **how did this tattoo artist come up with the idea of a quality tattoo ?** I would highly recommend this shop to anyone looking to get a quality tattoo done . |
| Multi | copy | It came out better than I even imagined . **. . this shop to anyone looking to get a quality tattoo done .** I would highly recommend this shop to anyone looking to get a quality tattoo done . |
| Mono | on-topic | Halbleiter-Riese National hatte die zwei Jahre zuvor akquirierte Chipschmiede Cyrix kürzlich an den taiwanischen Chipsatzproduzenten VIA weiter verkauft . **VIA ist die weltweit tätige Tochter von Cyrix ( GM ) .** 130-Nanometer-Chipfertigung läuft an |
| Multi | copy | Halbleiter-Riese National hatte die zwei Jahre zuvor akquirierte Chipschmiede Cyrix kürzlich an den taiwanischen Chipsatzproduzenten VIA weiter verkauft . **128 - Nanometer - Chipfertigung läuft an - an - an .** 130-Nanometer-Chipfertigung läuft an |
| Multi | on-topic | Keskuspankki sitoi Islannin kruunun kurssin euroon kaksi päivää sitten , jolloin eurolla sai 131 kruunua . **1900 - luvun alkuvuonna eurolla sai 140 kruunua .** Käytännössä tämä merkitsi vakavaa iskua Islannin taloudelle . |
| Multi | on-topic | I stadsmiljön utgör parker och grönområden en viktig del i våra dagar . **Stallar i naturen utgör också en viktig del i våra dagar .** Men naturen låg i omedelbar närhet , och stallar och ladugårdar var långt in på 1800-talet vanliga i den agrara svenska småstaden |

Figure 3: Example sentence generations (in bold) together with the manual quality judgements and the context provided in generation, for English, German, Finnish and Swedish.

For each language, we randomly sample 30 documents from the Universal Dependencies version 2.4 training data, and from each document we randomly select 2 sentences. For each of these sentences, we provide on input the preceding and following sentence as the left and right context for the model. Between these contexts, we use the parallel-sequential method of Wang and Cho to generate text which is as long, in terms of subword count, as the original sentence, restricting nevertheless to a minimum of 5 subwords and a maximum of 15 subwords. The maximum of 15 subwords was selected in preliminary experiments, as for considerably longer sequences, the model starts deviating from the seeded context and often fails to even stick to the language of the seed, owing to the fact that BERT is not trained to deal with long sequences of consecutive masked positions.

Subsequently, we manually evaluate the generated texts in context, and classify them into the following categories:

- **on-topic**: original, intelligible sentence or phrase without excessive errors, essentially fitting the context

- **off-topic**: original, intelligible sentence or phrase without excessive errors, not fitting the context

- **copy**: unoriginal text composed for the most part of verbatim copied sections of the context, often containing grammatical and flow errors

- **gibberish**: unintelligible sequence of words and characters, text with excessive grammatical and flow errors

The results of the evaluation are shown in Table 4. A comparison against an existing monolingual model is possible only for English and German. Both for English and German, there is a striking difference, where the monolingual models generate a substantially larger proportion of original on-topic text, compared to the multilingual model which, for the most part, copies sections of the context or produces gibberish. Especially for German, the monolingual model generates a subjectively very good output, with next to no copying and gibberish. For Finnish and Swedish, we can only report on the multilingual model, showing the same tendencies to copy or produce gibberish as for English and German. Overall, the results in Table 4 demonstrate that the multilingual model is clearly inferior to the monolingual counterparts and unsuitable for the generation task.

Figure 3 lists a few examples of generated sentences for the four languages and the available models. For English and German it illustrates the comparably worse performance of the multilingual model, as the generation is mostly copying from the context rather than creating original and fluent text that fits the context. For Finnish and Swedish, it shows cases where the generation has been able to fill in sentences that are correct and that to some extent relates to the context.

## 5 Discussion and Conclusions

In this paper, we set out to establish whether the multilingual BERT model, as distributed, is of sufficient quality to be considered an effective substitute for a dedicated, monolingual model for the given language. We tested the model on three tasks of increasing difficulty: a simple syntactic classification task, a cloze test, and full text generation. We found that the multilingual model notably lags behind the available monolingual models and the gap opens as the complexity of the task increases. While on the syntactic classification task, all models perform comparatively well, in the cloze test there is a notable difference. In the full text generation the multilingual model outputs are practically useless, while the monolingual models produce very good, and in the case of German rather impressive output. We can also observe major differences across languages in the multilingual model where, for instance, in the cloze test the model is considerably more likely to produce gibberish in Finnish than e.g. in German. It is not clear, however, to what extent this reflects the simple fact that Finnish has fewer "easy" functional words, providing for a harder task.

These results allow us to conclude that the current multilingual BERT model as distributed is not able to substitute a well-trained monolingual model in more challenging tasks. This, however, is unlikely due to the multilinguality of the model, rather, we believe it is due to the simple fact that each language is a mere 1/100th of the training data and training effort of the model. In other words, the model seems undertrained w.r.t. to individual languages. This is, for example, hinted at in the text generation task where the multilingual model mostly copies from the context or produces gibberish, while the monolingual models produce a considerably higher proportion of original text. Intuitively, this would fit a pattern where one would expect the model, as it is being trained, to first produce gibberish, then learn to understand and copy, and finally learn to generate.

The primary practical conclusion of this paper is that it is indeed necessary to invest the necessary computational effort to produce well-trained BERT models for other languages instead of relying on the present multilingual model as distributed. We also established baseline results on several tasks across several languages, allowing a better intuitive estimation of the applicability of the multilingual model in different situations.

## Acknowledgments

We gratefully acknowledge the support of the Google Digital News Innovation Fund, Academy of Finland, CSC – IT Center for Science, and the NVIDIA Corporation GPU Grant Program.